\documentclass[11pt]{article}
\usepackage[preprint]{acl}

\usepackage{times}
\usepackage{latexsym}
\usepackage[T1]{fontenc}
\usepackage[utf8]{inputenc}
\usepackage{microtype}
\usepackage{inconsolata}
\usepackage{graphicx}
\usepackage{booktabs}
\usepackage{multirow}
\usepackage{enumitem}
\usepackage{amsmath}
\usepackage{amssymb}
\usepackage{tikz}
\usetikzlibrary{shapes.geometric, arrows.meta, positioning, calc}

\graphicspath{{figures/}}

\newcommand{\abPoolCells}{28}
\newcommand{\abPoolMinMargin}{0.070}
\newcommand{\abPoolBestGain}{0.100--0.123}
\newcommand{\abLayerCost}{0.054--0.078}
\newcommand{\abSltRange}{0.632--0.703}
\newcommand{\abTbgRange}{0.522--0.524}
\newcommand{\abDepthRange}{42--50\%}
\newcommand{\abFullRange}{0.786--0.851}
\newcommand{\abFullTrainMax}{0.9995}
\newcommand{\abPcaGain}{0.028--0.102}

\newcommand{\abGapFull}{0.087--0.202}
\newcommand{\abGapPca}{0.009--0.013}
\newcommand{\abPcaNarrowMargin}{0.007}
\newcommand{\abPcaBandSpread}{0.007}
\newcommand{\abSupGain}{0.162--0.236}
\newcommand{\abSupGainMatched}{0.080--0.208}
\newcommand{\abSupRatioQwen}{28}
\newcommand{\abSupRatioLlama}{6.6}
\newcommand{\resBlendGain}{0.060}
\newcommand{\resBlendGainCI}{[0.050, 0.070]}
\newcommand{\resProbeGain}{0.030}
\newcommand{\resProbeGainCI}{[0.015, 0.045]}
\newcommand{\resGenMacro}{0.896}
\newcommand{\resGenMacroProbe}{0.857}
\newcommand{\resGenInflation}{0.028}
\newcommand{\resLuminaBlend}{0.865}
\newcommand{\resLuminaProbe}{0.818}
\newcommand{\resTaskMacroBlend}{0.901}
\newcommand{\resTaskMacroProbe}{0.859}

\newcommand{\resTaskBaseRates}{17.8--64.3\%}
\newcommand{\resProbeFone}{0.763}
\newcommand{\resStageOneFone}{0.792}
\newcommand{\resProbeFoneProprietary}{0.404--0.405}
\newcommand{\resBlendFoneProprietary}{0.563--0.579}
\newcommand{\resEceOverall}{0.073--0.088}
\newcommand{\resEceTaskMax}{0.121}
\newcommand{\alphaGrid}{21}
\newcommand{\alphaTuned}{0.9240}
\newcommand{\alphaUntuned}{0.9236}
\newcommand{\alphaGain}{0.0004}
\newcommand{\alphaFoneTuned}{0.820}
\newcommand{\alphaFoneUntuned}{0.804}
\newcommand{\alphaFoneGain}{0.016}
\newcommand{\alphaPairedCI}{[-0.0000, +0.0010]}
\newcommand{\selEvaluations}{944}
\newcommand{\selPairings}{8}
\newcommand{\selConfigs}{84}
\newcommand{\selHalves}{400}
\newcommand{\selOptimismAuroc}{0.002}
\newcommand{\selOptimismFone}{0.007}
\newcommand{\alphaHonestFone}{+0.012}

\newcommand{\selBestRecordedAuroc}{0.925}
\newcommand{\selProductRecomputed}{0.868}
\newcommand{\mcAuroc}{0.663}
\newcommand{\mcFone}{0.577}

\newcommand{\mcTaskRange}{0.490--0.605}

\newcommand{\mcProbeDelta}{0.216--0.231}
\newcommand{\mcProbeDeltaCI}{[+0.194, +0.254]}

\newcommand{\logoCtxSeeds}{5}
\newcommand{\logoCtxMacro}{-0.003}
\newcommand{\logoCtxCI}{[-0.018, +0.013]}

\newcommand{\logoSizeMacro}{+0.001}
\newcommand{\logoSizeCI}{[+0.000, +0.002]}
\newcommand{\logoNetMacro}{+0.001}
\newcommand{\logoNetCI}{[-0.011, +0.013]}
\newcommand{\logoBound}{0.02}
\newcommand{\logoDosePct}{46\%}
\newcommand{\logoNContexts}{2{,}514}

\newcommand{\logoMacroDelta}{+0.002 to +0.004}
\newcommand{\logoWorstDelta}{-0.012}
\newcommand{\logoBestDelta}{+0.027}
\newcommand{\cfLengthOnly}{0.664}
\newcommand{\cfLengthOnlySum}{0.515}
\newcommand{\cfLengthOnlyQa}{0.712}
\newcommand{\cfLengthOnlyDtt}{0.482}
\newcommand{\cfProbeWithinLength}{0.881}
\newcommand{\cfProbePooled}{0.894}
\newcommand{\alphaPlateauWorstAt}{0.30}
\newcommand{\alphaPlateauBestAt}{0.45}
\newcommand{\alphaPlateauFoneRange}{0.802--0.820}
\newcommand{\alphaPlateauSpread}{0.004}
\newcommand{\resProbeRange}{0.879--0.894}
\newcommand{\resProbeSpread}{0.015}
\newcommand{\resBlendAuroc}{0.924}
\newcommand{\resBlendFone}{0.820}
\newcommand{\resStageOneAuroc}{0.864}
\newcommand{\resTaskDetectorData}{0.897}
\newcommand{\resTaskDetectorQa}{0.855}
\newcommand{\resTaskDetectorSummary}{0.746}
\newcommand{\resTaskProbeData}{0.880}
\newcommand{\resTaskProbeQa}{0.858}
\newcommand{\resTaskProbeSummary}{0.838}
\newcommand{\resTaskProbeGainSummary}{+0.092}
\newcommand{\resGenBlendRange}{0.858--0.934}
\newcommand{\resGenBlendMin}{0.858}
\newcommand{\resGenBlendMax}{0.934}
\newcommand{\resBlendProprietary}{0.866}
\newcommand{\resBlendOpen}{0.913}
\newcommand{\resGenProbeSeventyB}{0.865}
\newcommand{\resGenDetectorSeventyB}{0.873}
\newcommand{\resGenCIMax}{0.13}
\newcommand{\resBlendFonePct}{82.0}
\newcommand{\resGenBlendLlamaThirteenB}{0.934}
\newcommand{\resProbeBestAuroc}{0.894}
\newcommand{\resProbeDataToText}{0.861--0.880}
\newcommand{\resProbeQa}{0.858--0.870}
\newcommand{\resProbeSummary}{0.770--0.838}
\newcommand{\resProbeFoneGapToOpt}{0.002--0.006}
\newcommand{\resEceStageOne}{0.101}
\newcommand{\resEceProbe}{0.073}
\newcommand{\resEceBlend}{0.037}
\newcommand{\resBlendThreeAuroc}{0.915}
\newcommand{\resBlendThreeFone}{0.809}
\newcommand{\cregGrid}{11}

\newcommand{\jdFone}{0.239}

\newcommand{\jdTaskWorst}{0.421}
\newcommand{\jdAllAuroc}{0.418--0.728}
\newcommand{\jdAllDelta}{+0.166 to +0.470}
\newcommand{\jdSmallestDelta}{+0.166}
\newcommand{\jdSmallestDeltaCI}{[+0.146, +0.186]}
\newcommand{\jdQwenAurocLow}{0.661}
\newcommand{\jdQwenAurocHigh}{0.728}
\newcommand{\jdLlamaAuroc}{0.418}
\newcommand{\dtDelta}{+0.006 to +0.010}
\newcommand{\dtDeltaMax}{+0.010}
\newcommand{\dtCIlow}{+0.001}
\newcommand{\dtCIhigh}{+0.015}
\newcommand{\dtFullTuned}{0.886--0.900}
\newcommand{\dtPcaDeployed}{0.879--0.894}
\newcommand{\dtFullDefault}{0.786--0.851}
\newcommand{\dtHardwareMax}{0.0000}
\newcommand{\dtHardwareR}{1.00000}
\newcommand{\dbBlendDelta}{+0.001 to +0.004}
\newcommand{\dbBlendDeltaMax}{+0.004}
\newcommand{\dbBlendHeadline}{+0.003}
\newcommand{\dbBlendHeadlineCI}{[-0.000, +0.006]}
\newcommand{\dbBlendHeadlineFone}{-0.011}
\newcommand{\dbBlendCIhigh}{+0.007}
\newcommand{\dbNullCount}{3}
\newcommand{\resTransferQa}{0.837--0.943}
\newcommand{\resTransferDialogue}{0.682--0.804}
\newcommand{\resTransferSummary}{0.643--0.704}
\newcommand{\resTransferSummaryLoss}{0.10--0.17}
\newcommand{\resLuminaDelta}{0.008}
\newcommand{\ccOverhead}{1.33}
\newcommand{\ccOverheadPct}{33}
\newcommand{\ccSlowdownDetector}{1.80}
\newcommand{\ccSlowdownObserver}{1.33}
\newcommand{\ccN}{300}
\newcommand{\latDetectorBase}{53}
\newcommand{\latDetectorLarge}{81}
\newcommand{\latObserverLow}{60}
\newcommand{\latObserverHigh}{198}
\newcommand{\latPooledLarge}{37}
\newcommand{\latPooledSmallest}{49}

\title{Grounding Probes: Generator-Independent Hallucination Detection\\
       from Observer Model Hidden States}

\author{Michael Rathmayr\textsuperscript{1} \and
  Ádám Kovács\textsuperscript{2} \and
  Gábor Recski\textsuperscript{2,1} \\
  \textsuperscript{1}TU Wien, Austria \quad \textsuperscript{2}KR Labs \\
  {\footnotesize
   \texttt{michael-rathmayr@gmx.at} \quad
   \texttt{kovacs@krlabs.eu} \quad
   \texttt{gabor.recski@tuwien.ac.at}}}

\begin{document}
\maketitle

\begin{abstract}
Detecting responses that retrieval-augmented generation does not ground in its context
trades speed against accuracy: surface checks miss paraphrased fabrication,
sampling-based methods cost extra generations.
Hidden-state probes sit between the two, but every existing one reads the generating
model's own activations, so a change of generator invalidates the detector and a
closed-weight generator is out of reach. This paper removes that coupling. The \emph{Grounding Probe} is
logistic regression over the mean-pooled middle-layer hidden states of an
\emph{observer} language model that reads the context, question, and response in one
forward pass and generates nothing, with the recipe it needs: pool over response tokens,
read a middle layer, and control capacity, which closes the train--test AUROC gap from
\abGapFull{} to \abGapPca{}. Asking the observer outright, rather than reading its
hidden state, costs at least \jdSmallestDelta{} AUROC in every one of four models.
Fitted on 15{,}090 annotated responses it reaches \resProbeRange{} AUROC on RAGTruth test
across four observers, and \resBlendAuroc{} AUROC with \resBlendFone{} F1@0.5 averaged with a supervised span
detector, \resBlendGain{} above that detector alone. One probe
holds across six generators, and hold-out controls, including one in which no evaluation
prompt appears in training, bound the cost of removing a generator at about \logoBound{}
AUROC. Code, probes, and predictions are released.
\end{abstract}

\section{Introduction}
\label{sec:intro}

Retrieval-augmented generation grounds a language model's output in retrieved
passages, and it does not stop the model from writing things the passages do not
support. On RAGTruth, GPT-4 produces unsupported content in 9.3\% of its
responses~\citep{ragtruth2024}, and weaker generators in the same benchmark
exceed 50\%. A production system must decide whether each response it is about to
return is grounded, and fast enough that the check fits inside the request.

Detection methods split along that latency budget. Supervised token-level
classifiers such as LettuceDetect~\citep{lettucedetect2025} annotate unsupported
spans in tens of milliseconds, but they work from surface evidence and can miss a
fluent paraphrase of something the source never said. Sampling-based methods catch
those cases: semantic entropy measures disagreement across several sampled
responses~\citep{farquhar2024semantic} at five to ten times the cost of generating
once~\citep{kossen2024seprobes}, seconds away from a serving path.

Hidden-state probes occupy the space between. A linear classifier on a
transformer's activations costs one forward pass and detects what those activations
already encode~\citep{alain2017understanding}, and applied to hallucination the
approach works: semantic entropy probes~\citep{kossen2024seprobes},
SAPLMA~\citep{azaria2023saplma}, HalluRAG~\citep{ridder2024hallurag},
ReDeEP~\citep{sun2025redeep}, and LUMINA~\citep{yeh2026lumina} all report usable
detection from model internals. They share an assumption: the activations come from
the model that generated the response. Changing the generator invalidates the
probe, and a generator behind a closed API cannot be probed at all, which excludes
the systems most likely to need the check.

\begin{figure}[t]
\centering
\resizebox{\columnwidth}{!}{%
\begin{tikzpicture}[
    node distance=0.45cm,
    box/.style={rectangle, draw, rounded corners=2pt, minimum height=0.55cm,
                text centered, font=\small, inner xsep=4pt},
    gen/.style={box, fill=black!5, draw=black!45},
    obs/.style={box, fill=blue!8, draw=blue!55, thick},
    prb/.style={box, fill=green!10, draw=green!45!black},
    arrow/.style={-{Stealth[length=2mm]}, thick},
    annot/.style={font=\scriptsize, text=black!60},
]
    \node[gen] (gen) {Any generator};
    \node[box, below=of gen] (triple) {(context, question, response)};
    \node[obs, below=of triple] (obs) {Observer LLM \; (frozen)};
    \node[box, below=of obs, align=center] (hs)
        {hidden states, layer $\ell$\\[-1pt]
         \scriptsize mean-pooled over response tokens};
    \node[prb, below=of hs] (probe) {z-score $\rightarrow$ PCA-512 $\rightarrow$ logistic regression};
    \node[box, below=of probe] (out) {$P(\text{hallucinated})$};

    \draw[arrow] (gen) -- (triple);
    \draw[arrow] (triple) -- (obs);
    \draw[arrow] (obs) -- (hs);
    \draw[arrow] (hs) -- (probe);
    \draw[arrow] (probe) -- (out);

    \node[annot, right=3pt of gen.east, anchor=west] {swappable};
    \node[annot, right=3pt of obs.east, anchor=west, align=left]
        {one forward pass,\\generates nothing};
    \node[annot, right=3pt of probe.east, anchor=west] {$d{+}1$ params};
\end{tikzpicture}%
}
\caption{The Grounding Probe. A frozen observer reads a response it did not
produce and generates no token; the probe reads its hidden states. One trained
probe serves any generator, including one behind a closed API.}
\label{fig:architecture}
\end{figure}
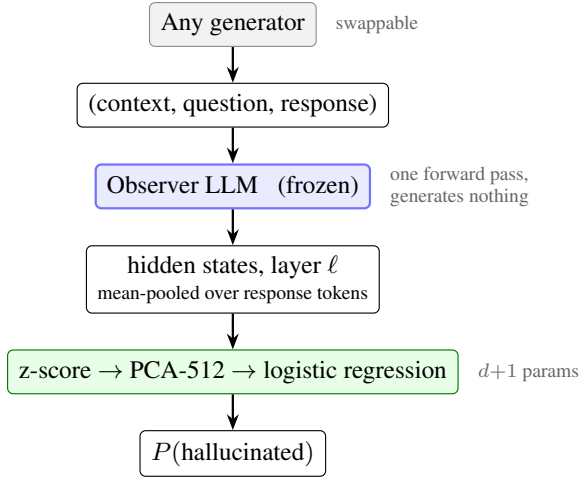

We drop the assumption. A separate model, which we call the \emph{observer}, reads
the finished triple of context, question, and response in one forward pass and
generates nothing; a linear probe on its mean-pooled middle-layer hidden states
outputs the probability that the response is unsupported
(Figure~\ref{fig:architecture}). The generator enters only through the text it
already produced, so one probe serves any generator, open or closed. Entailment
models such as MiniCheck~\citep{tang2024minicheck} and
AlignScore~\citep{zha2023alignscore} already evaluate responses with a separate
model, but they expose a scalar verdict; we read the representations that produced
it.

Three contributions follow.

\begin{enumerate}[nosep,leftmargin=*]
\item \textbf{Observer probing.} Hallucination detection from the hidden states of
a model that did not generate the response, which decouples the detector from the
generator. We evaluate one probe, fitted once, across six generators of the
RAGTruth test split.
\item \textbf{A recipe that generalizes from little data.} Mean pooling over
response tokens, a middle extraction layer, and a projection to 512 components.
Fifteen thousand annotated responses are enough, and without capacity control the
probe memorizes them. The projection supplies that control without tuning. Tuning the
regularization instead comes out marginally ahead on held-out data, inside the
\resProbeSpread{} AUROC margin we fixed in advance.
\item \textbf{Complementarity with token-level detection.} Averaging the probe
score with a supervised span detector's score improves on the detector on the
same test split, in both ranking and threshold behavior, without retraining it. We
release the code, the fitted probes, and the per-sample
predictions.\footnote{Code and per-sample predictions:
\url{https://github.com/MRathmayr/LettuceDetect}. Fitted probes:
\url{https://huggingface.co/mrathmayr/lettucedetect-grounding-probes}.}
\end{enumerate}

\section{Related Work}
\label{sec:related}

\paragraph{Detecting hallucination in RAG.}
RAGTruth~\citep{ragtruth2024} supplies word-level annotations over close to 18{,}000
responses from six generators and made supervised response-level detection
practical. LettuceDetect~\citep{lettucedetect2025} fine-tunes a long-context encoder
on those annotations to predict unsupported spans in one pass, and
Luna~\citep{luna2025} takes the same supervised route with a different backbone;
RAG-HAT~\citep{raghat2024} instead fine-tunes a generator to be
hallucination-aware. \citet{kovacs2026beyond} extend span-level detection to code and
tool output with a 2B generative detector; Section~\ref{sec:results-blend} says why we
blend with an encoder instead. A parallel line treats detection as entailment between source
and response, with purpose-built verifiers such as
MiniCheck~\citep{tang2024minicheck} or alignment models such as
AlignScore~\citep{zha2023alignscore} and SummaC~\citep{laban2022summac}. Both
families read only text, which makes them generator-independent and confines them
to what the surface shows.

\paragraph{Probing model internals.}
Linear probes were introduced to ask what a network's layers
encode~\citep{alain2017understanding}, with the caveat that a probe's success can
reflect its own capacity as much as the representation's
content~\citep{belinkov2022probing}. Applied to truthfulness the technique has been
productive. SAPLMA~\citep{azaria2023saplma} classifies statement truth from hidden
states; semantic entropy probes~\citep{kossen2024seprobes} predict a sampled
uncertainty measure from one pass, extracting at the second-last response token or the
last prompt token; HalluRAG~\citep{ridder2024hallurag} trains multi-layer classifiers
for RAG and reports a sharp cross-dataset drop. MIND~\citep{su2024mind} detects during
generation, HaloScope~\citep{du2024haloscope} without labels. ReDeEP~\citep{sun2025redeep} and LUMINA~\citep{yeh2026lumina} go further
into mechanism, scoring how much a response draws on retrieved context against
parametric knowledge, and report the strongest RAGTruth numbers among
internals-based methods.

Across that literature the probed model is the generator. Every method above
extracts activations from the model that produced the response under
evaluation, so a probe is valid for one generator and a closed-weight generator
is out of reach entirely. Entailment methods avoid the coupling by giving up
internal access. We are not aware of prior work that probes a separate evaluator
model which did not generate the response, and that combination is what this
paper reports.

\paragraph{Combining detectors.}
Routing inputs by difficulty is old and well understood, from boosted
cascades~\citep{viola2001cascade} to transformer early
exit~\citep{xin2020deebert,zhou2020pabee}. Hallucination detection has produced several
method families that could be combined this way, but they are typically evaluated in
isolation, on different splits and generators, so error correlation between families is
unmeasured. We report where averaging a response-level probe with a token-level detector
helps, and a routing architecture that underperforms that simpler alternative.

\section{Grounding Probes}
\label{sec:method}

A Grounding Probe is a logistic regression classifier over the hidden states of a
language model that reads a retrieval-augmented response without having produced it.
We call that model the \emph{observer}. It receives the finished triple of context,
question, and response as plain text, in the position of a reviewer reading a document
someone else wrote, and never emits a token. Generator independence follows from the
setup rather than the training data: the generator appears only through the text it
already produced.

\subsection{Observer setup}
\label{sec:method-observer}

The template concatenates the retrieved passages, the question, and a
\texttt{Response:} marker followed by the response. Prompt and response are tokenized
separately, with a leading space on the response so subword tokenization at the boundary
matches training. Inputs over 4096 tokens are truncated from the right. Nothing here is
tuned.

The observer then runs one forward pass with hidden-state output enabled and
generation disabled. That pass is the entire inference cost. Because the observer
is frozen and consumes only text, one probe scores responses from any generator,
including one behind a closed API whose internals are unavailable in principle.

Generator coupling is replaced by two narrower couplings. The inference template must
match the one used to fit the probe, and the observer must be loaded in the training
precision, fp16 here. We did not evaluate the released probes under quantization.

\subsection{Hidden state extraction}
\label{sec:method-extraction}

We read the hidden states at one layer $\ell$, keep only response-token positions, and
mean-pool them into $\mathbf{h} \in \mathbb{R}^{d}$, with $d$ from 2048 for Qwen2.5 3B to
5120 for 14B, cast to float32. Dropping the context and question positions is deliberate:
responses drawing on the same passage share those tokens whether or not they are faithful,
so the signal must live in how the observer represents the response.
Section~\ref{sec:ablations} settles the layer.

\subsection{Classifier}
\label{sec:method-classifier}

Three fitted transformations turn $\mathbf{h}$ into a probability. Z-score
normalization centers and scales each dimension using training-set statistics
$\boldsymbol{\mu}$ and $\boldsymbol{\sigma}$:
\begin{equation}
\tilde{\mathbf{h}} = \frac{\mathbf{h} - \boldsymbol{\mu}}{\boldsymbol{\sigma}}
\label{eq:zscore}
\end{equation}
This step matters because L2 regularization penalizes every weight equally, so
without it the high-variance dimensions are effectively under-regularized
regardless of how useful they are. A principal component projection then reduces
the normalized vector to its leading $k = 512$ components:
\begin{equation}
\mathbf{x} = \mathbf{W}_{\mathrm{PCA}}^{\top} \tilde{\mathbf{h}},
\qquad \mathbf{W}_{\mathrm{PCA}} \in \mathbb{R}^{d \times k}
\label{eq:pca}
\end{equation}
Logistic regression maps the result to the hallucination probability:
\begin{equation}
P(\text{hallucinated} \mid \mathbf{x}) =
\sigma(\mathbf{w}^{\top}\mathbf{x} + b)
\label{eq:sigmoid}
\end{equation}
The three stages compose into one. Writing
$\mathbf{S} = \operatorname{diag}(\boldsymbol{\sigma})$, the deployed probe is a single
affine functional of the raw hidden state,
$\sigma(\tilde{\mathbf{w}}^{\top}\mathbf{h} + \tilde{b})$ with
$\tilde{\mathbf{w}} = \mathbf{S}^{-1}\mathbf{W}_{\mathrm{PCA}}\mathbf{w}$, so inference
is one dot product in $\mathbb{R}^{d}$, under 0.5\,ms, and the observer forward pass
dominates. That also locates what the projection does: it adds nothing at inference and
confines $\tilde{\mathbf{w}}$ to the span of the leading $k$ components while the probe
is fitted, leaving $k+1$ free parameters rather than $d+1$. The three stages are fitted
together and serialized as one 4 to 11\,MB artifact.

Linearity is deliberate. A linear probe recovers only what the
hidden states already encode linearly~\citep{alain2017understanding}, whereas a
multi-layer probe can compute the target itself and blurs what the representation
carries into what the probe invents. Training minimizes binary cross-entropy with
L2 regularization at $C = 1.0$ using L-BFGS, with \texttt{class\_weight} set to
\texttt{balanced}. The projection and $C$ turn out to be substitutes rather than
complementary, tuning $C$ at full dimensionality landing marginally ahead, so we deploy
the projection at the default $C$, which needs no search. Section~\ref{sec:ablations} sweeps both.

\subsection{Training}
\label{sec:method-training}

Probes are fitted on the RAGTruth training split, 15{,}090 responses with human
span-level annotations~\citep{ragtruth2024}, collapsed to one binary label per
response: $y = 1$ if any span is unsupported. RAGTruth ships the responses, so no
generation happens at training time either, and fitting reduces to one forward pass
per sample plus a convex optimization, 14 to 50 minutes per observer on an A100. The
probe predicts $P(\text{hallucinated})$ directly, with no score inversion.

\subsection{Score blending}
\label{sec:method-blending}

A response-level probe and a token-level detector see different things, so we also
report their combination. Given the probe score $s_3$ and a token-level span score
$s_1$ max-pooled to the response level, the blend is
\begin{equation}
s = \alpha s_1 + (1 - \alpha) s_3
\label{eq:blend}
\end{equation}
with $\alpha = 0.45$ throughout, applied to every sample. We chose blending over
confidence routing, which forwards only uncertain samples and replaces the
token-level score, because routing costs F1 for structural reasons rather than
through a bad threshold choice (Section~\ref{sec:routing}).

\begin{table*}[t]
\centering
\small
\begin{tabular}{lcccc}
\toprule
Method & AUROC & 95\% CI & F1@0.5 & Latency (ms) \\
\midrule
\multicolumn{5}{l}{\emph{Prior work on RAGTruth (per-generator protocol)}} \\
\quad SAPLMA \citep{azaria2023saplma} & 0.803 & -- & -- & -- \\
\quad SEP \citep{kossen2024seprobes} & 0.809 & -- & -- & -- \\
\quad ReDeEP \citep{sun2025redeep} & 0.824 & -- & -- & -- \\
\quad LUMINA \citep{yeh2026lumina} & 0.857 & -- & -- & -- \\
\quad LettuceDetect-v2-2B \citep{kovacs2026beyond} & -- & -- & 0.818 & -- \\
\midrule
\multicolumn{5}{l}{\emph{Measured here on the same 2700 rows (mixed-generator protocol)}} \\
\quad MiniCheck-Flan-T5-L~\citep{tang2024minicheck} & 0.663 & [0.642, 0.683] & 0.577 & 131$^{\dagger}$ \\
\quad Prompted judge (Qwen2.5 3B) & 0.661 & [0.640, 0.682] & 0.239 & 41$^{\ddagger}$ \\
\quad LettuceDetect-base~\citep{lettucedetect2025} & 0.840 & -- & 0.751 & 53 \\
\quad LettuceDetect-large~\citep{lettucedetect2025} & 0.864 & -- & 0.792 & 81 \\
\quad Grounding Probe (Qwen2.5 3B) & 0.879 & [0.865, 0.892] & 0.738 & 60 \\
\quad Grounding Probe (Qwen2.5 7B) & 0.884 & [0.871, 0.898] & 0.752 & 119 \\
\quad Grounding Probe (Llama 3.1 8B) & 0.888 & [0.874, 0.901] & 0.755 & 120 \\
\quad Grounding Probe (Qwen2.5 14B) & 0.894 & [0.881, 0.907] & 0.763 & 198 \\
\quad \, + LettuceDetect-large (Qwen2.5 3B) & 0.915 & -- & 0.809 & 81 \\
\quad \, + LettuceDetect-large (Qwen2.5 14B) & \textbf{0.924} & [0.913, 0.935] & \textbf{0.820} & 198 \\
\bottomrule
\end{tabular}
\caption{Response-level detection on the RAGTruth test split ($n=2700$), with all
six generators pooled for the rows in the lower block. The first three prior-work rows are as reported
by \citet{sun2025redeep} on single-generator LLaMA outputs, and LUMINA reports its own
figure evaluating GPT-4 and ChatGPT separately, so the blocks are not comparable; the matched
comparison is per generator (Table~\ref{tab:generator},
Section~\ref{sec:results-generator}). MiniCheck is a verdict-style entailment
verifier we ran ourselves on these rows, so it is comparable to the rows below it;
$^{\dagger}$its latency was measured on a GTX 1080 and is not comparable to the
A100 column. The prompted judge is the same observer as the 3B probe row, at the
same precision, so those two differ only in how the answer is read;
$^{\ddagger}$its latency comes from a separate rented A100 and is likewise not
comparable with the column.
\citet{kovacs2026beyond} report example-level F1 only, hence the empty AUROC cell.
Latency is otherwise a measured per-sample mean on an A100 over these same
2700 rows, for a probe row the observer forward pass, and for a blend the per-sample
$\max(t_{\mathrm{S1}}, t_{\mathrm{S3}})$, which measured concurrent execution
exceeds by 1.33$\times$, so a blend entry is a floor rather than an estimate
(Section~\ref{sec:latency}). CIs are
percentile bootstrap over 10{,}000 resamples.}
\label{tab:main}
\end{table*}

\section{Experimental Setup}
\label{sec:setup}

RAGTruth~\citep{ragtruth2024} is the primary benchmark. Its training split
(15{,}090 responses) fits the probes and its held-out test split (2700
responses, 34.9\% hallucinated) carries every result below unless stated
otherwise. The test split is balanced across three task types at 900 responses
each, question answering, data-to-text generation, and summarization, and across the
six generators named in Table~\ref{tab:generator} at 450 responses each.
HaluEval~\citep{li2023halueval} supplies
30{,}000 further samples, used only to measure transfer.

Four observers supply hidden states: Qwen2.5 3B, 7B, and 14B~\citep{qwen2.5} and
Llama 3.1 8B~\citep{llama3}, read at layers $-15$, $-12$, $-20$, and $-16$
respectively and projected to 512 components. LettuceDetect~\citep{lettucedetect2025}
provides the token-level baseline in both released sizes, base (150M parameters)
and large (396M), and is also the codebase our implementation extends. All
measurements run on an A100 40\,GB. Numeric precision is matched between probe
fitting and inference, fp16 in both cases, because mismatching it moves the
hidden-state distribution and we measured AUROC drops of 0.02--0.16 when it was
mismatched by accident during development.

AUROC is the primary metric, being threshold-free across methods whose scores live on
different scales. For threshold-dependent performance we report F1 at the fixed
default of 0.5 rather than a threshold tuned on the test split, which would be an
upper bound. Intervals are percentile bootstrap over 10{,}000 resamples, paired on
samples where the quantity is a difference.

Three decisions need stating up front, two selections and one reporting population. The blending weight $\alpha$ was
selected over a \alphaGrid{}-point grid computed on the evaluation split, not on
held-out data, and every configuration we report uses that single value. The
exposure is small and measurable: at $\alpha = 0.50$, the equal-weight default
requiring no selection at all, the headline configuration reaches \alphaUntuned{}
AUROC against \alphaTuned{} at $\alpha = 0.45$, so selection is worth
\alphaGain{} AUROC, with a paired interval on that difference of \alphaPairedCI{}.
Section~\ref{sec:routing} has the full sweep. F1@0.5 is the sensitive quantity,
\alphaFoneTuned{} against \alphaFoneUntuned{}, because shifting $\alpha$ moves the
score distribution underneath a fixed threshold. Read \alphaFoneGain{} as an in-sample gain
over one particular default rather than as a bound: $\alpha = 0.50$ is not the plateau's
worst point, which sits at \alphaPlateauWorstAt{}, and the out-of-sample version of the
same choice, made on one random half of the split, is worth \alphaHonestFone{} F1 on the
other. Fitting $\alpha$ on the training
data would not repair any of this, since both components are themselves fitted on
RAGTruth train; an honest treatment needs a third split and a fresh extraction run.

The headline is a selected configuration in a wider sense too: the best of
\selPairings{} detector--observer pairings, on a split that has carried about
\selEvaluations{} scored configurations. Choosing among the \selConfigs{} combinations of
$\alpha$ and observer on one random half and scoring the other, over \selHalves{} halves,
the in-sample maximum overstates held-out performance by \selOptimismAuroc{} AUROC and
\selOptimismFone{} F1@0.5. Intervals below are conditional on that selection.

The projection width is the same kind of decision: PCA-512 was chosen on held-out test
AUROC and every number we report uses it. Its margin over PCA-256 is at most
\abPcaNarrowMargin{} AUROC on paired intervals that overlap zero for three of the four
observers, so we present capacity control as necessary and the width as not
(Section~\ref{sec:ablations}). The extraction layer was chosen by cross-validation on
training data alone.

The third decision concerns the latency population. Every latency we report is a
per-sample mean over the 2700 evaluation rows, drawn from a 12{,}700-sample benchmark
run whose other inputs are shorter HaluEval samples that carry no figure in
Table~\ref{tab:main}. The
restriction reorders the cheap pairing: pooled, the smallest observer averages
\latPooledSmallest{}\,ms against \latPooledLarge{}\,ms for the large detector, and on
the evaluation population that ordering reverses, \latObserverLow{} against
\latDetectorLarge{}\,ms. Section~\ref{sec:latency} derives the blend figures.

\section{Results}
\label{sec:results}

\subsection{Grounding probes alone}
\label{sec:results-probe}

A linear probe on observer hidden states reaches \resProbeRange{} AUROC on the RAGTruth
test split across the four observers (Table~\ref{tab:main}). Performance tracks observer
size, but the spread over a 4.7-fold parameter range is only \resProbeSpread{}, untested pairwise, so
a 3B observer captures most of the signal.

All three task types are detectable: data-to-text at \resProbeDataToText{}, question
answering at \resProbeQa{}, summarization lowest at \resProbeSummary{}. Summarization is
lowest for every observer; the data-to-text and question-answering gap is within noise
and reverses for Qwen2.5 3B. RAGTruth summarization compresses the longest contexts into
the shortest responses, but length does not explain the gap: it ranks responses at
\cfLengthOnlySum{} on summarization and at \cfLengthOnlyDtt{} on data-to-text, which the
probe finds no harder, against \cfLengthOnlyQa{} on question answering.

At the pooled base rate the probe outputs usable probabilities and not only a usable
ranking: F1 at the default threshold of 0.5 falls within \resProbeFoneGapToOpt{} of the best F1 any
threshold achieves here, and expected calibration error stays in \resEceOverall{}. That
is what makes the score safe to average with another detector's, and it depends on the
supervision (Section~\ref{sec:ablations}).

Two verdict-style evaluators on the same rows test the central claim directly.
MiniCheck~\citep{tang2024minicheck} returns a scalar judgment on the same triple,
reaching \mcAuroc{} AUROC and \mcFone{} F1@0.5, which puts the four probes
\mcProbeDelta{} above it, paired interval \mcProbeDeltaCI{}; being a 770M Flan-T5, it
changes the model as well as the readout. Prompting the observer removes that confound:
asking each of the four in the probe's fp16, from the same template, whether any part
of the response is unsupported by the context and scoring $P(\text{Yes})$, which points
the same way as the probe's $P(\text{hallucinated})$, gives \jdAllAuroc{} AUROC against
\resProbeRange{} for the probe read from the same model, a paired \jdAllDelta{} with
every interval excluding zero, narrowest \jdSmallestDelta{} \jdSmallestDeltaCI{}. Judge
quality climbs with observer size across the Qwen family, \jdQwenAurocLow{} to
\jdQwenAurocHigh{}, without approaching the probe; on Llama 3.1 8B the same template
lands below chance at \jdLlamaAuroc{}. Both readouts cost one
forward pass, so this is not a speed argument: the hidden state carries what the decoded
verdict does not. Pooling flatters both, spanning \mcTaskRange{} by task for MiniCheck
and falling to \jdTaskWorst{} on data-to-text for the 3B judge, whose F1@0.5 of
\jdFone{} appears to follow from an almost complete refusal to answer Yes.

Two qualifications belong with those numbers. The probe's F1@0.5 of \resProbeFone{} is
below the detector's \resStageOneFone{}, so it wins on ranking and loses at the default
operating point, which is one reason we combine the two. The calibration holds at the
pooled 34.9\% hallucination rate and degrades away from it: per-task calibration error
reaches \resEceTaskMax{}, and on the two generators near a 10\% rate the probe's F1@0.5
falls to \resProbeFoneProprietary{}.

\subsection{Combining with a token-level detector}
\label{sec:results-blend}

Blending the probe with LettuceDetect-large lifts AUROC from \resStageOneAuroc{} to \resBlendAuroc{}
and F1@0.5 from \resStageOneFone{} to \resBlendFone{} on the identical split
(Table~\ref{tab:main}). The gain
over the detector alone is \resBlendGain{} AUROC, and because both methods score the
same responses we can put a paired bootstrap interval on the difference itself:
\resBlendGainCI{}. The F1 gain is the more selection-dependent of the two: at the
untuned $\alpha = 0.50$ it is \alphaFoneUntuned{} rather than \alphaFoneTuned{},
so \alphaFoneGain{} of the \alphaFoneTuned{} figure comes from choosing $\alpha$. The probe alone contributes \resProbeGain{},
interval \resProbeGainCI{}, over the detector, and reaches \resProbeBestAuroc{}, so neither component
explains the combination.

Per task the two cover different ground. The detector scores \resTaskDetectorData{} on data-to-text,
where surface overlap with structured records is informative, \resTaskDetectorQa{} on question
answering and \resTaskDetectorSummary{} on summarization; the probe scores
\resTaskProbeData{}, \resTaskProbeQa{} and \resTaskProbeSummary{}, so its
advantage is largest where the detector is weakest, $\resTaskProbeGainSummary{}$ on summarization.

Calibration improves with accuracy: expected calibration error on equal-width bins is
\resEceStageOne{} for the detector, \resEceProbe{} for the probe and \resEceBlend{} for
the blend, so averaging scores that err in different directions moves the result toward
observed frequencies. Nor does the gain need a large observer, the same detector with
Qwen2.5 3B giving \resBlendThreeAuroc{} AUROC and \resBlendThreeFone{} F1@0.5 at the
lowest observer latency we measure, \latObserverLow{}\,ms.

Two published RAGTruth F1 figures sit at or above ours, and we claim to beat neither:
RAG-HAT~\citep{raghat2024} reports 83.9 against our \resBlendFonePct{}, and the 2B
generative detector of \citet{kovacs2026beyond} reports 81.8, which our
\resBlendFonePct{} matches rather than exceeds. We
keep the encoder detector as the blending partner because the blend needs a continuous
response-level score, which a token classifier gives directly and a model emitting spans
as generated text does not, and because the encoder in that newer family scores below
the one we use on RAGTruth.

\subsection{Independence from the generator}
\label{sec:results-generator}

\begin{table*}[t]
\centering
\small
\begin{tabular}{lcccccc}
\toprule
Generator & $n$ & Hall.\ rate (\%) & LettuceDetect-large & Grounding Probe & Blend & 95\% CI \\
\midrule
GPT-4 & 450 & 9.3 & 0.767 & 0.811 & 0.872 & [0.807, 0.930] \\
GPT-3.5-turbo & 450 & 10.2 & 0.731 & 0.824 & 0.858 & [0.789, 0.918] \\
Mistral-7B-Instruct$^{\dagger}$ & 450 & 55.8 & 0.882 & 0.886 & 0.931 & [0.906, 0.953] \\
Llama-2-7B-chat$^{\dagger}$ & 450 & 50.2 & 0.820 & 0.842 & 0.875 & [0.841, 0.906] \\
Llama-2-13B-chat$^{\dagger}$ & 450 & 46.0 & 0.885 & 0.916 & 0.934 & [0.908, 0.957] \\
Llama-2-70B-chat$^{\dagger}$ & 450 & 38.0 & 0.873 & 0.865 & 0.904 & [0.871, 0.934] \\
\midrule
Macro-average & 450 & 34.9 & 0.826 & 0.857 & 0.896 & -- \\
Pooled (mixed-generator) & 2700 & 34.9 & 0.864 & 0.894 & 0.924 & [0.913, 0.935] \\
\bottomrule
\end{tabular}
\caption{Detection AUROC by the generator that produced the response.
$^{\dagger}$ marks open-weight generators, 1800 of the 2700 test responses. One
probe, fitted once, is applied to all six with no per-generator tuning. The pooled
figure exceeds the macro-average by +0.028 because
hallucination rates range from
9.3\% to
55.8\%, so pooling adds base-rate
separation and makes the task easier, not harder. Comparisons against
per-generator literature must use the matched generator subset
(Section~\ref{sec:results-generator}).}
\label{tab:generator}
\end{table*}

We break the test split out by the model that produced each response
(Table~\ref{tab:generator}). RAGTruth carries six generators at 450 test responses each,
four of them open-weight, and the training split carries the same six, so these are
in-distribution generators unless a probe is refitted with one held out. The predictions
do not record the generator, so we recover it from the positional sample identifiers and
accept the join only when label and task type agree on all 2700 rows, which they do; that
check is necessary rather than sufficient.

One probe, fitted once and never tuned per generator, scores \resGenBlendRange{} blended
across the six. The weakest is GPT-3.5-turbo at \resGenBlendMin{} and the strongest Llama-2-13B-chat at
\resGenBlendMax{}, on
450 responses each, so we do not read the spread as an ordering. Grouped by provenance
the blend reaches \resBlendProprietary{} pooled over the two proprietary generators and
\resBlendOpen{} pooled over the four open-weight ones, but that grouping tracks hallucination rate rather than provenance.
The probe does not win everywhere; on Llama-2-70B-chat it scores \resGenProbeSeventyB{} against the
detector's \resGenDetectorSeventyB{}.

Comparing this against per-generator numbers in the literature needs care. Our pooled
figure, \resBlendAuroc{}, is \emph{higher} than the macro-average over the six generators,
\resGenMacro{}, by \resGenInflation{}: hallucination rates run from 9.3\% to 55.8\%,
so pooling lets base-rate differences contribute to the ranking and makes the task
easier, not harder. The same holds for task type, whose rates span
\resTaskBaseRates{}, giving a task macro-average of \resTaskMacroBlend{} against
\resBlendAuroc{} blended and \resTaskMacroProbe{} against \resProbeBestAuroc{} for the probe.

Because all six generators appear in probe training, per-generator evaluation alone
cannot speak to an unseen one. We refit with one held out under three designs, a positive delta meaning the held-out
probe scored higher. Splitting a
generator's rows at matched training size moves macro AUROC by \logoMacroDelta{} across
the four observers, and per generator by \logoWorstDelta{} to \logoBestDelta{}, but that
leaves two things uncontrolled. RAGTruth pairs each of its \logoNContexts{} contexts with
all six generators, so a held-out generator's evaluation prompts stay in training under
the other five; splitting on context removes them and gives \logoCtxMacro{}, averaged
over \logoCtxSeeds{} seeds, with \logoCtxCI{} across the six generators. Matching
training size also displaces other generators' rows, so a size-only arm keeps the
generator in and measures that displacement alone at \logoSizeMacro{} \logoSizeCI{};
subtracting it within that arm leaves \logoNetMacro{} \logoNetCI{} for generator
presence. Read these as bounds. The row and the context design put zero inside intervals
of half-width near \logoBound{} AUROC, and that width is variation between generators
rather than between splits. The size-only arm is the exception: its effect is small but
detectable, the interval excluding zero below the three decimals shown.

A matched comparison has to name the population.
ReDeEP~\citep{sun2025redeep} reports 0.824 on Llama-2-13B outputs, where our blend
scores \resGenBlendLlamaThirteenB{}, though our blend consumes RAGTruth's train-split annotations twice
over, through the detector and the probe, while ReDeEP's mechanistic score does not.
LUMINA~\citep{yeh2026lumina} reports 0.857 evaluating GPT-4 and
ChatGPT separately, and macro-averaged over exactly those two generators our blend reaches
\resLuminaBlend{}, a difference of \resLuminaDelta{} that sits well inside the per-generator confidence
intervals. Against LUMINA's protocol, then, we do not claim an improvement. The probe on
its own is weaker still on that protocol, \resLuminaProbe{} over LUMINA's two generators
and \resGenMacroProbe{} macro-averaged over all six, so what the method offers against
LUMINA is not accuracy but a detector that does not need the generator's weights.

\section{Analysis}
\label{sec:analysis}

\begin{table}[tb]
\centering
\footnotesize
\setlength{\tabcolsep}{2.0pt}
\begin{tabular}{@{}lcccc@{}}
\toprule
 & \multicolumn{3}{c}{Qwen2.5} & Llama \\
\cmidrule(lr){2-4} \cmidrule(lr){5-5}
Configuration & 3B & 7B & 14B & 8B \\
\midrule
\multicolumn{5}{l}{\emph{Token aggregation and layer} (5-fold CV, 5{,}000 train)} \\
\quad Mean pool, selected layer & \textbf{0.733} & \textbf{0.778} & \textbf{0.803} & \textbf{0.785} \\
\quad Mean pool, final layer & 0.668 & 0.700 & 0.736 & 0.732 \\
\quad Second-last token (SLT) & 0.632 & 0.655 & 0.703 & 0.685 \\
\quad Prompt-final token (TBG) & 0.524 & 0.523 & 0.522 & 0.523 \\
\midrule
\multicolumn{5}{l}{\emph{Dimensionality} (RAGTruth test)} \\
\quad Full ($d = 2048$--$5120$) & 0.851 & 0.810 & 0.801 & 0.786 \\
\quad PCA-256 & 0.877 & 0.884 & 0.892 & 0.881 \\
\quad PCA-512 & \textbf{0.879} & \textbf{0.884} & \textbf{0.894} & \textbf{0.888} \\
\quad PCA-1024 & 0.874 & 0.884 & 0.892 & 0.882 \\
\midrule
\multicolumn{5}{l}{\emph{Training supervision} (RAGTruth test)} \\
\quad QA correctness, no context & 0.642 & 0.723 & 0.662 & 0.706 \\
\quad Hallucination labels + context & \textbf{0.879} & \textbf{0.884} & \textbf{0.894} & \textbf{0.888} \\
\bottomrule
\end{tabular}
\caption{Ablations, AUROC. The first block is 5-fold cross-validation on 5{,}000
training samples, comparable only within the block. SLT and TBG are the positions
used by \citet{kossen2024seprobes}, each at its own best layer, which favours them;
mean pooling still wins by 0.100--0.123. ``Selected layer'' is the
deployed layer ($-15$ / $-12$ / $-20$ / $-16$ for 3B/7B/14B/8B), chosen by this sweep. The supervision
block changes the training data: 428{,}619 QA samples scored by answer-overlap F1
without context (100{,}000 for Llama 3.1 8B) against 15{,}090 RAGTruth samples with
human labels and context. Those QA-supervised probes are also full-dimensional, at
their own layer, and 4-bit, so the row is not single-factor; against the
full-dimensional row above the gap narrows to 0.080--0.208.}
\label{tab:ablation}
\end{table}

\subsection{What the probe needs}
\label{sec:ablations}

Four choices decide whether an observer probe works at all
(Table~\ref{tab:ablation}). None of them touches the classifier.

\paragraph{Pool over the response.}
Mean pooling across response-token positions beats reading a single position at
every one of the \abPoolCells{} layer and observer combinations we swept. The
smallest margin is \abPoolMinMargin{} AUROC, and \abPoolBestGain{} when each method
gets its own best layer. \citet{kossen2024seprobes} extract at the second-last
response token or the last prompt token and report late layers as best; both are
far weaker here. The pre-generation token barely separates the classes at
\abTbgRange{}, near chance, which is expected since that position precedes the
response being judged. The second-last token carries real signal, \abSltRange{},
and pooling still adds a tenth or more on top.

\paragraph{Read from the middle, not the end.}
The best layer sits at \abDepthRange{} of the stack counted back from the output,
just past the midpoint: $-15$ of 36 layers for Qwen2.5 3B, $-12$ of 28 for 7B,
$-20$ of 48 for 14B, $-16$ of 32 for Llama 3.1 8B. Reading the final layer instead
costs \abLayerCost{} under the same protocol. The sweep covered the upper half of each
stack in seven steps, so a better layer further down is not ruled out, and for Llama
3.1 8B the selected layer is the deepest tested.

\begin{figure*}[t]
\centering
\includegraphics[width=\textwidth]{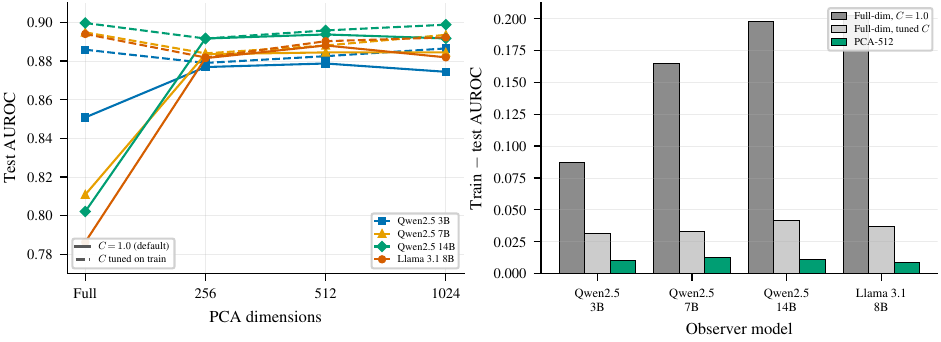}
\caption{Capacity control, not dimensionality. Left: test AUROC against the number of
components, at the default $C = 1.0$ and at a $C$ cross-validated on the training
split. The default-$C$ curve climbs steeply out of full dimensionality and peaks at or
near 512 components, which is what makes the projection look necessary; the tuned-$C$
curve is flat, with full dimensionality slightly ahead except at the smallest observer. Right: the train--test gap,
which PCA-512 collapses from \abGapFull{} at the default $C$ to \abGapPca{}. Tuning $C$
closes most of the same gap without projecting, the middle bar.
Section~\ref{sec:ablations} reads the two as substitutes and keeps the projection,
which needs no search.}
\label{fig:pca}
\end{figure*}

\paragraph{Project before fitting.}
A full-dimensional probe fits its 15{,}090 training samples almost perfectly, up to
\abFullTrainMax{} training AUROC for Qwen2.5 14B, and reaches only \abFullRange{} on
test. Reducing to 512 components adds \abPcaGain{} test AUROC and collapses the
train--test gap from \abGapFull{} to \abGapPca{} (Figure~\ref{fig:pca}). Between 256
and 1024 components test AUROC moves by at most \abPcaBandSpread{}.

Projection is not the only route to that. With the L2 strength swept over \cregGrid{}
values at both widths, $C$ selected on the training split alone and test scored once, a
full-dimensional probe reaches \dtFullTuned{} against \dtPcaDeployed{} for the deployed
PCA-512: \dtDelta{} per observer, every paired interval excluding zero and the four
spanning \dtCIlow{} to \dtCIhigh{}. The same states at the default $C$ give
\dtFullDefault{}, so regularization alone recovers what the projection was credited
with, and PCA-512 barely responds to $C$. We fixed the decision rule before those states
existed: move the deployed recipe only if the held-out margin beat \resProbeSpread{},
the spread across observer sizes (Section~\ref{sec:results-probe}) that this paper
already treats as minor. At \dtDeltaMax{} it does not, so PCA-512 stays deployed for
needing no search, not for being the better probe. Capacity control rather than
dimensionality is what the sweep isolates, now on held-out data. Extraction hardware
does not enter: on those test states, taken from a rented A100, the deployed probe
agrees with the cluster extraction to \dtHardwareMax{} AUROC at per-sample
$r = \dtHardwareR{}$.

Swapping that probe into the blend would not repay the search either. Through the same
blending pipeline at the same $\alpha$, blended AUROC moves \dbBlendDelta{} across the
observers, at most \dbBlendDeltaMax{}, and \dbBlendHeadline{} \dbBlendHeadlineCI{} at
the headline pairing, \dbNullCount{} of the four intervals containing zero, so the
evidence excludes a blend gain larger than \dbBlendCIhigh{} AUROC in any observer.
Fixed-threshold F1 at 14B moves the other way, \dbBlendHeadlineFone{}, as strong
regularization pulling probabilities toward 0.5 would predict.

\paragraph{Match the supervision to the deployment.}
Holding the classifier and the observers fixed, 15{,}090 RAGTruth samples with human
labels and the retrieved context beat question-answering samples labeled by
answer-overlap F1 without context, by \abSupGain{} AUROC. That comparison is not
single-factor: the QA-supervised probes are also full-dimensional, read at their own
swept layer, and loaded with 4-bit weights. Against the full-dimensional grounding row,
which removes the largest of those differences, the gap narrows to \abSupGainMatched{}
and stays positive for all four observers, and the mismatched set is larger by
\abSupRatioLlama$\times$ to \abSupRatioQwen$\times$. For a specific
deployment, labeled examples in its own format beat more of them in another.

\subsection{Cross-dataset transfer}
\label{sec:transfer}

\begin{figure}[t]
\centering
\includegraphics[width=\columnwidth]{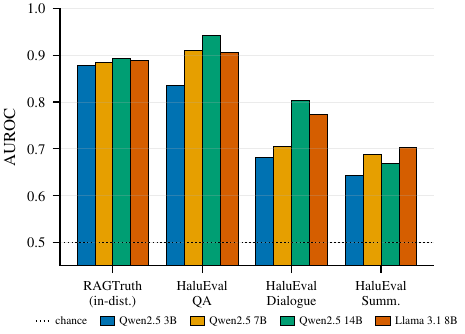}
\caption{Transfer to HaluEval for probes trained only on the RAGTruth training
split. Question
answering holds up (\resTransferQa{}); dialogue and summarization degrade but stay
well above chance. RAGTruth contains no dialogue data, which is the most
plausible reason that split is weakest.}
\label{fig:transfer}
\end{figure}

Probes fitted only on the RAGTruth training split transfer unevenly to HaluEval
(Figure~\ref{fig:transfer}), with no HaluEval training data: question answering holds up
at \resTransferQa{}, above the in-domain figure for three of four observers, while
dialogue drops to \resTransferDialogue{} and summarization to \resTransferSummary{}. RAGTruth contains no dialogue,
the most plausible reason that split is weakest, and that does not extend to
summarization, which RAGTruth covers and which still loses \resTransferSummaryLoss{} AUROC against its
RAGTruth counterpart. What transfers is narrower than a shared task label suggests.

\subsection{Blending weight, and why not routing}
\label{sec:routing}

The blend is insensitive to $\alpha$ over a wide interval. Across
$\alpha \in [0.30, 0.50]$ AUROC varies by at most \alphaPlateauSpread{} for every
configuration, while F1@0.5 traces a hump over \alphaPlateauFoneRange{} that peaks at
$\alpha = \alphaPlateauBestAt{}$, since $\alpha$ slides the score distribution
underneath a fixed cut. A deployment
can start from equal weights and lose almost nothing in ranking quality; the
threshold is what needs attention.

Confidence routing looks like the natural architecture and performs worse. Routing only
uncertain samples to the probe, replacing the token-level score, raises AUROC over the
large detector by 0.019--0.029 while costing 0.026--0.042 in F1 at the best threshold
(OptF1), because no single cut
suits the bimodal mixture routing produces. Averaging every sample removes the problem;
Appendix~\ref{sec:appendix} has the mechanism and the caveat that the two architectures
do not consume identical inputs.

\subsection{Latency}
\label{sec:latency}

\begin{figure}[t]
\centering
\includegraphics[width=\columnwidth]{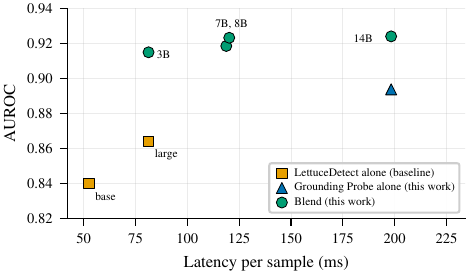}
\caption{Accuracy against latency. Per-stage latencies are measured per sample on an
A100 over the evaluation rows; blend latency is the per-sample
$\max(t_{\mathrm{S1}}, t_{\mathrm{S3}})$,
standing in for concurrent execution on separate devices and excluding coordination
overhead. Measured concurrent execution costs \ccOverhead{}$\times$ that maximum on a
two-A100 host (Section~\ref{sec:latency}). Semantic entropy is not
plotted: at seconds per query it sits one to two orders of magnitude to the right, on a
different benchmark and setup~\citep{farquhar2024semantic,kossen2024seprobes}.}
\label{fig:pareto}
\end{figure}

Measured per sample on an A100 over the evaluation rows, the token-level detector takes
\latDetectorBase{}\,ms in its base size and \latDetectorLarge{}\,ms in its large size,
and an observer forward pass \latObserverLow{} to \latObserverHigh{}\,ms
(Table~\ref{tab:main}), so observer size moves the total further than detector size
does. For the blend we report the per-sample
$\max(t_{\mathrm{S1}}, t_{\mathrm{S3}})$, what concurrent execution on separate devices
would cost before coordination overhead: the large detector's \latDetectorLarge{}\,ms
in the cheap pairing, and the 14B observer's \latObserverHigh{}\,ms in the accurate one.
Sequential execution costs the sum. That substitution is optimistic. Running the two
stages at once on two A100s in one machine, each pinned to its own card, over \ccN{}
samples, costs \ccOverhead{}$\times$ the derived maximum, \ccOverheadPct{}\% above what
the table reports, with the span detector slowed \ccSlowdownDetector{}$\times$ and the
observer \ccSlowdownObserver{}$\times$. Separate cards put the contention on the host, and we
did not attribute it further. Table~\ref{tab:main} keeps the derived
figure, its column coming from one run while the overhead is a ratio from another
machine. All of it stays one
to two orders of magnitude below the seconds per query that sampling-based detection
costs~\citep{farquhar2024semantic,kossen2024seprobes}
(Figure~\ref{fig:pareto}).

\section{Conclusion}
\label{sec:conclusion}

Hallucination detection from hidden states does not require the generator's. An observer
that merely reads a retrieval-augmented response encodes whether it is grounded, linearly
enough to reach \resProbeRange{} AUROC, \resBlendAuroc{} blended, and to beat what that model says when
asked outright.

\section*{Limitations}

We removed the probe's dependence on the generator and replaced it with a
dependence on the observer. Each probe is tied to one observer model, one
extraction layer, and one numeric precision, and a probe fitted on one
observer's hidden states cannot be applied to another's. Swapping observers means
re-extracting hidden states, under an hour per model on an A100, plus a
one-time layer sweep for an unfamiliar architecture family. Whether a probe
transfers across observers is untested. We expect it does not, since the two
hidden state spaces are unaligned.

Our evidence rests largely on one benchmark. The probes are fitted on RAGTruth's
training split and evaluated on its disjoint test split, but one annotation convention
governs both, so test performance cannot by itself show that a probe tracks
groundedness rather than that convention. Transfer to HaluEval question answering at \resTransferQa{}
argues against the narrowest version of that concern, and the weaker dialogue and
summarization transfer argues for keeping it open. RAGTruth is also public and
predates all four observers, so its responses may have appeared in their
pretraining data.

Our generator-independence evidence is a per-generator evaluation plus three hold-out
controls on the training side (Section~\ref{sec:results-generator}), each run for every
observer on the stored train-split hidden states. Those states are too large to release,
so the refits are not verifiable from our artifacts, though we do release the per-sample
probabilities they produced. The controls bound the cost of removing a generator at
roughly \logoBound{} AUROC rather than showing the cost is zero, they measure the
probe alone rather than the blend we headline, and the matched arm trains on only about
\logoDosePct{} of a generator's deployed row count. All six generators date from 2023, and the split contains 450 responses per generator, so the
per-generator confidence intervals are wide, spanning up to \resGenCIMax{} AUROC. On the
lowest-prevalence generator only 42 of 450 responses are hallucinated, where a
percentile bootstrap may under-cover.

The blend latency we report is derived, a per-sample $\max$ over two separately
measured stages, with the sequential figure their sum. That substitution has now been
tested and it is optimistic: concurrent execution costs \ccOverhead{}$\times$ the
derived maximum (Section~\ref{sec:latency}). Only the ratio carries over, that run
being a different machine, so the blend column is a floor rather than an estimate.

The verdict comparison rests on one prompt. We wrote a single zero-shot template, in
the probe's input format, with no calibration and no tuning per model family, and that
is enough to account for a judge below chance: an untuned prompt can collapse the
verdict distribution of a model it was not written for, which is what \jdLlamaAuroc{}
on Llama 3.1 8B looks like. We do not know the cause. Every judge figure here is a
lower bound on what prompting that observer could reach, and the \jdSmallestDelta{}
minimum gap is measured against that bound. Nor do we report a
probe on a generator's own hidden states under this protocol, which would measure what
dropping generator access costs, or a non-linear probe, which would test the word
``linearly''; the former is obstructed for the two closed-weight generators.

Two selection decisions ran on the evaluation split: the blending weight, worth
\alphaGain{} AUROC and \alphaFoneGain{} F1@0.5 over the untuned default, and the
projection width, worth at most \abPcaNarrowMargin{} AUROC over the next setting.
Both exposures are bounded and neither is zero.

Pooled evaluation flatters every headline number here. RAGTruth's hallucination rates
differ across both generators and task types, so pooling contributes base-rate
separation on top of groundedness: the blend's task macro-average is
\resTaskMacroBlend{} against \resBlendAuroc{} pooled and the probe's is
\resTaskMacroProbe{} against \resProbeBestAuroc{}. Response length is one correlate we can bound. Length alone ranks
responses at \cfLengthOnly{} AUROC, and restricting the probe to compare within
deciles of response length moves it from \cfProbePooled{} to
\cfProbeWithinLength{}, so length accounts for little of the probe's ranking. Other
correlates of groundedness we have not isolated.

Threshold behavior is weaker than the pooled figures suggest. The probe's F1@0.5
of \resProbeFone{} is below the detector's \resStageOneFone{}, per-task calibration error
reaches \resEceTaskMax{}, and on the two generators with roughly 10\% hallucination
rates the probe's F1@0.5 drops to \resProbeFoneProprietary{} and the blend's to
\resBlendFoneProprietary{}. Under transfer the default threshold is further off
still. A deployment whose base rate differs from RAGTruth's must set its own
threshold, and our transfer numbers are reported as AUROC only.

Summarization remains the weakest task type at \resProbeSummary{}, and it is the task
where a detector is most useful, since summarization hallucinations are the least
visible to surface checks. The longest RAGTruth test input reaches roughly two thirds of the 4096-token budget
and none is truncated, so the long-context regime is untested. Everything reported here is English, and we
make no claim about other languages or about specialized domains such as medicine,
law, and finance.

\section*{Acknowledgments}

The computational results have been achieved using the Austrian Scientific
Computing (ASC) infrastructure. The training-split hidden states, the fitted
probes, and the layer and pooling sweeps reported here were computed on the
Vienna Scientific Cluster (VSC).

\bibliography{references}

\appendix
\section{Blending and Routing Detail}
\label{sec:appendix}

For both detector sizes paired with the smallest and the largest observer, every
pairing improves on its detector (Table~\ref{tab:routing}). The base detector
gains more than the large one, since it starts further back.

Routing's F1 cost has a structural cause. Samples resolved at the token level keep
scores near zero while escalated samples get probe scores spread across the middle, so
the routed score is a bimodal mixture and no single threshold suits it. A sweep over 72
threshold configurations moves AUROC by less than 0.004, which is what one expects if
the score scales rather than the cut points are the problem. One asymmetry is worth
flagging: routing scores on a composite that adds lexical overlap and static embedding
features to the token-level detector, so routing and blending do not consume identical
inputs.

Two hyperparameters in this comparison were selected on evaluation data rather
than on a held-out split, and neither can be re-selected without a fresh
extraction run. The blending weight $\alpha = 0.45$ comes from an \alphaGrid{}-point grid
over evaluation predictions; Section~\ref{sec:setup} quantifies what that
selection is worth. The weighted average is also not the highest AUROC our sweep
recorded. A product rule appears at \selBestRecordedAuroc{}, above the blend's
\alphaTuned{}, but we cannot reproduce that value from the stored per-sample
scores, which give \selProductRecomputed{}, and its F1@0.5 is below the blend's
either way, so we report the weighted average and flag the discrepancy. The routing confidence thresholds, $t_{\mathrm{low}} = 0.09$
and $t_{\mathrm{high}} = 0.79$, were calibrated on 500 samples drawn from the
same test split. At those values the first stage resolves about 54\% of samples
without escalating, measured over the full 12{,}700-sample benchmark and on the
composite score the cascade routes on, which adds lexical overlap and static
embedding features to the token-level detector. Since routing loses to blending on F1 by a margin far larger
than any threshold choice moves it (Section~\ref{sec:routing}), the selection
does not affect the conclusion we draw from the comparison.

\begin{table}[t]
\centering
\small
\begin{tabular}{lccc}
\toprule
Configuration & Detector & Blend & F1@0.5 \\
\midrule
LD-base + Qwen 3B & 0.840 & 0.899 & 0.776 \\
LD-base + Qwen 14B & 0.840 & 0.910 & 0.791 \\
LD-large + Qwen 3B & 0.864 & 0.915 & 0.809 \\
LD-large + Qwen 14B & 0.864 & 0.924 & 0.820 \\
\bottomrule
\end{tabular}
\caption{Score blending at $\alpha = 0.45$ across LettuceDetect (LD) base/large
and two observer sizes, RAGTruth test AUROC. The ``Detector'' column is the
token-level model alone. Every pairing improves on its token-level detector, and
the ordering of the pairings is unchanged at the untuned $\alpha = 0.50$.
Confidence routing, which forwards only uncertain samples and replaces rather
than combines the score, raises AUROC over the detector alone but costs
0.026--0.042 OptF1 for the large detector.}
\label{tab:routing}
\end{table}

\end{document}